\documentclass{article}

\PassOptionsToPackage{numbers, compress}{natbib}

\usepackage[preprint]{neurips_2024}

\usepackage[utf8]{inputenc} 
\usepackage[T1]{fontenc}    
\usepackage{hyperref}       
\usepackage{url}            
\usepackage{booktabs}       
\usepackage{amsfonts}       
\usepackage{nicefrac}       
\usepackage{microtype}      
\usepackage{xcolor}         
\usepackage{amsmath} 
\usepackage{graphicx}

\usepackage{amssymb}  
\usepackage{colortbl}
\usepackage{times}
\usepackage{epsfig}
\usepackage{soul}
\usepackage{caption}
\usepackage{subcaption}
\usepackage{bbold}
\usepackage{wrapfig,lipsum}

\usepackage{multirow}
\usepackage{float}
\usepackage{tabulary}
\usepackage{multicol} 
\usepackage{pifont}   

\title{VSSD:  Vision Mamba with \\ Non-Causal State Space Duality}

\author{%
	Yuheng Shi  \\
	City University of Hong Kong \\
	\texttt{yuhengshi99@gmail.com} \\
	\And
	Minjing Dong\\
	City University of Hong Kong\\
	\texttt{minjdong@cityu.edu.hk}\\
	\And
	Mingjia Li\\
	Tianjin University\\
	\texttt{mingjiali@tju.edu.cn}\\
	\And
	Chang Xu\\
	University of Sydney\\
	c.xu@sydney.edu.au \\
}

\definecolor{sasacolor}{HTML}{000000}
\definecolor{vitcolor}{HTML}{fc8e62}
\definecolor{convcolor}{HTML}{412F8A}
\definecolor{redcolor}{HTML}{fc6562}
\definecolor{pa}{HTML}{C209C1}

\definecolor{citationcolor}{HTML}{007ED2}

\newcommand{\grayrow}{\rowcolor[gray]{.95}}

\DeclareMathAlphabet\mathbfcal{OMS}{cmsy}{b}{n}

\newcommand{\xmark}{\ding{55}}

\begin{document}

	\maketitle

	\begin{abstract}
		Vision transformers have significantly advanced the field of computer vision, offering robust modeling capabilities and global receptive field. However, their high computational demands limit their applicability in processing long sequences. 
		To tackle this issue, State Space Models (SSMs) have gained prominence in vision tasks as they offer linear computational complexity. Recently, State Space Duality (SSD), an improved variant of SSMs, was introduced in Mamba2 to enhance model performance and efficiency. 
		However, the inherent causal nature of SSD/SSMs restricts their applications in non-causal vision tasks. 
		To address this limitation, we introduce Visual State Space Duality (VSSD) model, which has a non-causal format of SSD. Specifically, we propose to discard the magnitude of interactions between the hidden state and tokens while preserving their relative weights, which relieves the dependencies of token contribution on previous tokens. Together with the involvement of multi-scan strategies, we show that the scanning results can be integrated to achieve non-causality, which not only improves the performance of SSD in vision tasks but also enhances its efficiency. We conduct extensive experiments on various benchmarks including image classification, detection, and segmentation, where VSSD surpasses existing state-of-the-art SSM-based models. Code and weights are available at \url{https://github.com/YuHengsss/VSSD}.
		
	\end{abstract}

	\section{Introduction}
	
	In recent years, vision transformers ~\cite{ViT16x16,Swin,Swinv2,guo2022cmt,PVT,nat,cswin,DeiT2021}, pioneered by the Vision Transformer (ViT)~\cite{ViT16x16}, have achieved tremendous success in the field of computer vision. Thanks to the global receptive field and the robust information modeling capabilities of the attention mechanism, models~\cite{Swin, zhang2022dino,cheng2021mask2former} based on vision transformers have advanced significantly in various tasks such as classification~\cite{deng2009imagenet}, detection~\cite{coco}, and segmentation~\cite{ade20k}, surpassing classic CNN-based models~\cite{simonyan2014very,resnet,senet,densenet}. However, the quadratic computational complexity of the attention mechanism makes it resource-intensive for tasks involving long sequences, which limits its broader application.
	
\begin{figure}
		\centering
		\includegraphics[width=1.0\linewidth]{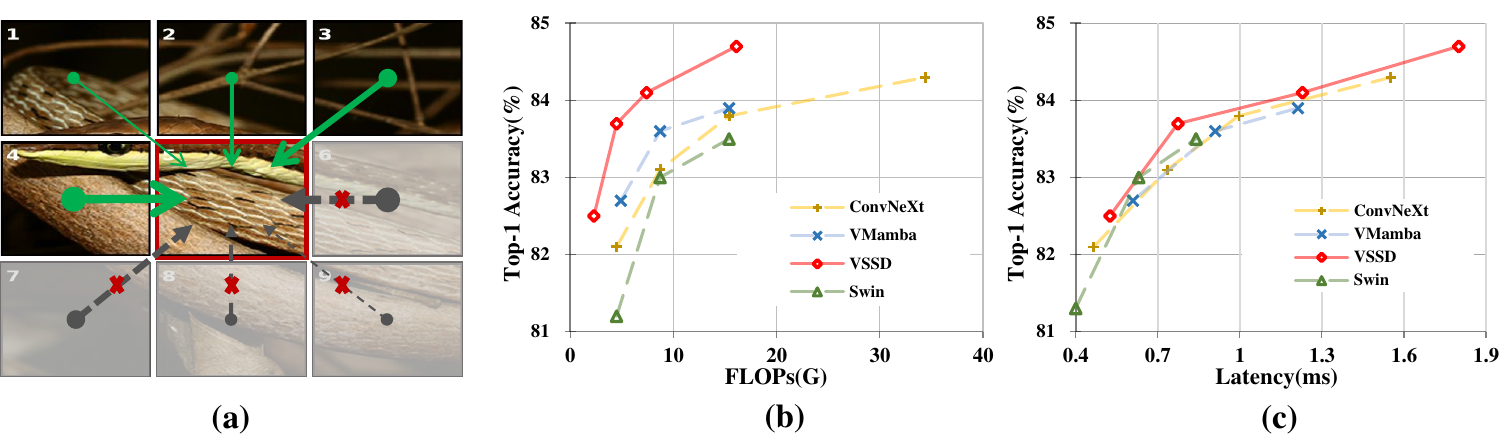}
		\caption{(a) Two challenges when applying SSM/SSD to image data. (b) and (c) are comparisons on ImageNet. Our VSSD model achieves leading accuracy and efficiency compared to CNN-based ConvNeXt~\cite{convnext}, ViT-based Swin Transformer~\cite{Swin}, and SSM-based VMamba~\cite{liu2024vmamba}. The latency of all models is measured on an A100 GPU using a batch size of 128 and FP16 precision.}
		\label{fig:comparision}
		\vspace{-3mm}
	\end{figure}

	Recently, State Space Models (SSMs)~\cite{gu2020hippo,gu2021lssl,gu2021s4,smith2022s5}, exemplified by Mamba~\cite{gu2023mamba}, have garnered considerable attention from researchers. The S6 block, in particular, offers a global receptive field and exhibits linear complexity with respect to sequence length, presenting an efficient alternative. 
	Pioneering vision mamba models such as Vim~\cite{zhu2024ViM} and VMamba~\cite{liu2024vmamba} have been developed to apply SSMs to vision tasks. Afterward, many variants were proposed~\cite{huang2024localmamba,efficientnet,yang2024plainmamba,shi2024multi}, which flatten 2D feature maps into 1D sequences using different scanning routes, model them with the S6 block, and subsequently integrate the results in multiple scanning routes. 
	These multi-scan approaches improve the performance of SSMs in vision tasks, achieving results competitive with those of both CNN-based and ViT-based methods.
	More recently, Mamba2~\cite{dao2024transformers} has introduced further enhancements to the S6 block, proposing the concept of State Space Duality (SSD). Mamba2 treats the state space transition matrix $\mathbf{A}$ as a scalar and expands the state space dimension, thereby enhancing model performance as well as training and inference efficiency. 
	However, there exists a major concern regarding the application of SSD/SSMs in vision tasks, where the image data is naturally non-causal while SSD/SSMs have inherent causal properties.  While another concern is flattening 2D feature maps into 1D sequences disrupts the inherent structural relationships among patches. 
	We provide an illustration in Fig.~\ref{fig:comparision} (a) to facilitate a more intuitive understanding of these two concerns.
	In this example, the central token within the flattened 1D sequences is restricted to accessing only previous tokens, unable to integrate information from subsequent tokens. Additionally, the token 1, which is adjacent to the central token in the 2D space, becomes distantly positioned in the 1D sequence, disrupting the natural structural relationships.
	A common practice in previous solutions \cite{liu2024vmamba,huang2024localmamba} is to increase the scanning routes on non-causal visual features, which alleviates these two concerns to some extent. Given these observations, an important question emerges: Is there a more effective and efficient way to apply SSD to non-causal vision data compared to multi-scan methods?
	
	To address this question, our analysis of SSD reveals that treating the matrix $\mathbf{A}$ as a scalar facilitates a straightforward transformation of the SSD into a non-causal and position-independent manner, which we denote as Non-Causal SSD (NC-SSD). 
	Specifically, rather than using $\mathbf{A}$ to determine the proportion of the hidden state to be retained, we employ it to dictate the extent of the current token's contribution to the hidden states. 
	In this scenario, each token's contribution becomes self-referential. Based on this property, we demonstrate that the causal mask in SSD can be naturally removed without the need for specific scanning routes. This observation motivates us to develop a non-causal format of SSD, in which a single global hidden state can be derived to replace the previous token-wise hidden states, resulting in not only improved accuracy but also enhanced training and inference speeds.
	Unlike previous multi-scan methods~\cite{zhu2024ViM,liu2024vmamba} that primarily alleviate the causal limitations of SSMs, our proposed NC-SSD also resolves the issue where flattening 2D feature maps into 1D sequences disrupts the continuity of adjacent tokens.
	Besides the NC-SSD, other techniques including hybrid with standard self-attention and overlapped downsampling are also explored. 
	Building on these techniques, we introduce our Visual State Space Duality (VSSD) model and demonstrate its superior effectiveness and efficiency relative to methods based on CNNs, ViTs, and SSMs, as illustrated in Fig.~\ref{fig:comparision} (b) and (c).
	Concretely, compared to the recent proposed SSM-based VMamba~\cite{liu2024vmamba}, our VSSD model outperforms it by approximately 1\% in top-1 accuracy on ImageNet-1K dataset~\cite{deng2009imagenet} while keeping a similar computational cost. Additionally, our model also consistently leads in the accuracy-latency curve.
	Besides the better trade-offs between performance and efficiency, another highlight of VSSD lies in the training speed. For example, compared to the vanilla SSD or multi-scan SSD (e.g., Bi-SSD with bidirectional scanning), our proposed model accelerates training speed by nearly 20$\%$ and 50$\%$ respectively.
	
	In summary, our contributions are twofold. 
	First, we analyze the state space duality and demonstrate that it can be seamlessly converted to a non-causal mode. 
	Based on this insight, we introduce the NC-SSD, which retains the global receptive field and linear complexity benefits of the original SSD while incorporating an inherent non-causal property and achieving improved training and inference efficiency. 
	Second, utilizing NC-SSD as the foundational component, we propose the VSSD model and conduct extensive experiments to validate its effectiveness. With similar parameters and computational costs, our VSSD model outperforms other State-Of-The-Art(SOTA) SSM-based models across several widely recognized benchmarks in classification, object detection, and segmentation.

	\section{Related Work}
	\textbf{Vision Transformers.} The introduction of Vision Transformers (ViTs)~\cite{ViT16x16,Swin,PVT,cswin,DeiT2021} has revitalized the field of computer vision, which was previously dominated by Convolutional Neural Networks (CNNs)~\cite{krizhevsky2012imagenet,simonyan2014very,resnet,xie2017aggregated,densenet,mobilenet,efficientnet,convnext}. However, the quadratic computational complexity of the self-attention mechanism in ViTs poses significant challenges when processing high-resolution images, demanding significant computational resources. To address this issue, different solutions have been proposed, including hierarchical architectures~\cite{Swin,Swinv2,cswin,PVT,pvtv2,tnt}, windowed attention~\cite{Swin,nat,tu2022maxvit,biformer}, and variants of self-attention~\cite{internimage,xia2023dat++,poolformer}. Meanwhile, linear attentions~\cite{katharopoulos2020transformers,choromanski2020rethinking,qin2022cosformer,han2024demystify} have been successful in reducing the computational complexity to a linear scale by changing the computation order of query, key and value in self-attention. Despite this advancement, the performance of linear attention remains inferior to that of the quadratic self-attention~\cite{attention} and its variants~\cite{nat,fan2024rmt,biformer}.
	
	\textbf{State Space Models.} 
	State Space Models (SSMs)~\cite{gu2020hippo,gu2021lssl,gu2021s4,smith2022s5,fu2022h3,gu2023mamba} have increasingly captured the attention of researchers due to their global receptive field and linear computational complexity. Mamba~\cite{gu2023mamba}, a prominent example of SSMs, introduced the S6 block, achieving performance on par with or better than transformers in Nature Language Processing (NLP) benchmarks. Subsequent efforts~\cite{pei2024efficientvmamba,huang2024localmamba,du2024understanding,yang2024plainmamba,chen2024mim,li2024videomamba,yang2024remamber,ruan2024vm} have explored the adaptation of the S6 block to vision tasks, yielding competitive results compared to both CNNs and ViT-based models. A central challenge in developing Mamba-based vision models is adapting the inherently causal properties of the Mamba block for non-causal image data. The most direct approach involves using different scanning routes to flatten 2D feature maps into 1D sequences which are then modeled with the S6 block and integrated.
	Inspired by these considerations, various scanning routes have been employed and proven effective, as evidenced by multiple studies~\cite{zhu2024ViM, liu2024vmamba, huang2024localmamba, pei2024efficientvmamba, shi2024multi}.
	More recently, Mamba2~\cite{dao2024transformers} has highlighted the significant overlap between state space models and structured masked attention, identifying them as duals of each other, and introduced the concept of State Space Duality (SSD). Building on this foundation, we demonstrate that the SSD can be transformed into a non-causal mode through a straightforward transformation, without the need of specific scanning route.
	
	\section{Method}
	\subsection{Preliminaries}
	\textbf{State Space Models.} Classic State Space Models (SSMs) are used to describe the dynamics of a continuous system, transforming an input sequence \(x(t) \in \mathbb{R}\) to a latent space representation \(h(t) \in \mathbb{R}^{N}\). This representation is then utilized to generate an output sequence \(y(t) \in \mathbb{R}\). The mathematical formulation of an SSM is structured as follows:
	\begin{equation}\label{eq:continuous_ssm}
		\begin{array}{l}
			h'(t) = \overset{\scriptscriptstyle\circ}{\mathbf{A}}  h(t) + \overset{\scriptscriptstyle\circ}{\mathbf{B}}  x(t),\;
			y(t)  = \mathbf{C}h(t),
		\end{array}
	\end{equation}
	where $\overset{\scriptscriptstyle\circ}{\mathbf{A}}  \in \mathbb{R}^{N\times N}$, $\overset{\scriptscriptstyle\circ}{\mathbf{B}}  \in \mathbb{R}^{N \times 1}$ and $\mathbf{C}\in \mathbb{R}^{1 \times N}$ are parameters.
	To effectively integrate continuous SSMs into deep learning architectures, discretization is essential. This process involves introducing a timescale parameter $\mathbf{\Delta} \in \mathbb{R}$ and applying the zero-order hold (ZOH) technique for discretization. Through this approach, the continuous matrices $\overset{\scriptscriptstyle\circ}{\mathbf{A}}$ and $\overset{\scriptscriptstyle\circ}{\mathbf{B}}$ are transformed into their discrete counterparts, $\mathbf{A}$ and $\mathbf{B}$. Consequently, Eq.~\ref{eq:continuous_ssm} is redefined in a discrete format in Eq.~\ref{eq:dis_ssm}, facilitating its application within modern computational frameworks:
	\begin{equation}\label{eq:dis_ssm}
		\begin{array}{l}
			h(t) = \mathbf{A}h(t-1) + \mathbf{B}x(t),\;
			y(t)  = \mathbf{C}h(t), \\
			\text{where }\mathbf{A} = e^{\mathbf{\Delta} \overset{\scriptscriptstyle\circ}{\mathbf{A}}},\;
			\mathbf{B} = (\mathbf{\Delta} \overset{\scriptscriptstyle\circ}{\mathbf{A}})^{-1}( e^{\mathbf{\Delta} \overset{\scriptscriptstyle\circ}{\mathbf{A}}} - \mathbf{I})\mathbf{\Delta} \overset{\scriptscriptstyle\circ}{\mathbf{B}}\approx \mathbf{\Delta} \overset{\scriptscriptstyle\circ}{\mathbf{B}},\\ 
		\end{array}
	\end{equation}
	where $\mathbf{I}$ denotes the identity matrix. Furthermore, the process of Eq.~\ref{eq:dis_ssm} could be implemented in a global convolution manner as:
	\begin{equation}
		\begin{array}{l}
			y  =x \odot \mathbf{K}, \; 
			\mathbf{K}  =\left(\mathbf{C} \mathbf{B}, \mathbf{C} \mathbf{A B}, \ldots, \mathbf{C}  \mathbf{A}^{L-1} \mathbf{B}\right),
		\end{array}
		\label{eq:conv_kernel}
	\end{equation}
	where $\mathbf{K} \in \mathbb{R}^{L}$ represents the convolution kernel. Recently, Mamba~\cite{gu2023mamba} makes the parameters $\mathbf{B}, \mathbf{C}$, and $\mathbf{\Delta}$ input-dependent. This modification addresses the limitations of the Linear Time Invariant (LTI) characteristics inherent in previous SSM models~\cite{gu2021s4,fu2022h3}, thus enhancing the adaptability and performance of SSMs.
	
	\subsection{Non-Causal State Space Duality}
	\label{sec:nc_ssd}
	More recently, Mamba2~\cite{dao2024transformers} introduced the State Space Duality (SSD) and simplified the matrix $\mathbf{A}$ into a scalar. 
	This special case of selective State Space Models (SSMs) can be implemented in both linear and quadratic forms. Without loss of generality, the matrix transformation form of selective state space models is expressed as follows:
	\begin{equation}
		\begin{array}{l}
			y(t) =\sum_{i=1}^{t} \mathbf{C}_{t}^{T} \mathbf{A}_{t: i+1} \mathbf{B}_{i} x(i), \text{where }\mathbf{A}_{t: i} = \prod_{i=2}^{t} \mathbf{A}_{i},  \\
			y =\operatorname{SSM}(\mathbf{A}, \mathbf{B}, \mathbf{C})(x)=\mathbf{F} x,\text{where } \mathbf{F}_{j i}=\mathbf{C}_{j}^{T} \mathbf{A}_{j: i} \mathbf{B}_{i}.  \\
		\end{array}
		\label{eq:mat_ssm}
	\end{equation}
	When $\mathbf{A}_i$ is reduced to a scalar, the quadratic form of Eq.~\ref{eq:mat_ssm} can be reformulated as:
	\begin{equation}
		y =\mathbf{F} x = \mathbf{M} \cdot (\mathbf{C}^{T}\mathbf{B})x,
		\text{ where } \mathbf{M}_{i j}=\left\{\begin{array}{ll}
			A_{i+1} \times \cdots \times A_{j} & i > j \\
			1 & i=j \\
			0 & i<j,
		\end{array}\right.
		\label{eq:mat_ssm_reformulated}
	\end{equation}
	while its linear form is denoted as:
	\begin{equation}
		\begin{array}{l}
			h(t) = A_t h(t-1) + \mathbf{B}_t x(t), 
			y(t) = \mathbf{C}_t h(t).
			\label{eq:ssd_linear}
		\end{array}
	\end{equation}
	To adapt SSMs for image data, the 2D feature maps should first be flattened into a 1D sequence of tokens which are then sequentially processed. Due to the causal nature of SSMs, where each token can only access previous tokens, information propagation is inherently unidirectional. This causal property leads to suboptimal performance when handling non-causal image data, a finding that has been corroborated by previous works~\cite{zhu2024ViM,liu2024vmamba,yu2024mambaout}. Furthermore, flattening the 2D feature maps into 1D sequences disrupts their intrinsic structural information. For instance, tokens that are adjacent in the 2D map might end up being far apart in the 1D sequence, resulting in a loss of performance in vision tasks~\cite{guo2024mambair}. Since SSD is a variant of SSMs, adopting SSD for vision tasks presents similar challenges to those observed with SSMs:
	\begin{itemize}
		\item \textbf{Challenge 1:} The causal property of the model restricts the flow of information, preventing later tokens from influencing earlier ones.
		
		\item \textbf{Challenge 2:} Flattening 2D feature maps into 1D sequences disrupts the inherent structural relationships among patches during processing.
	\end{itemize}

	In the context of applying causal SSD to non-causal image data, it is instructive to revisit the linear formulation of SSD. 
	In Eq.~\ref{eq:ssd_linear}, the scalar $A_t$ modulates the influence of the previous hidden state $h(t-1)$ and the information in current time step. 
	In other words, the current hidden state $h(t)$ could be viewed as a linear combination of the previous hidden state and the current input, weighted by $A_t$ and 1, respectively. 
	Therefore, if we discard the magnitude of these two terms and only preserve their relative weight, Eq.~\ref{eq:ssd_linear} could be rewritten as: 
	\begin{equation}
		h(t) = h(t-1) + \frac{1}{A_t}  \mathbf{B}_t x(t) = \sum_{i=1}^{t} \frac{1}{A_i} \mathbf{B}_i x(i).
		\label{eq:ssd_linear_rewritten}
	\end{equation}
	In this scenario, the contribution of a specific token to the current hidden state can be directly determined by itself as $\frac{1}{A_i}$, rather than through the cumulative multiplication of multiple coefficients. 
	With each token's contribution becoming self-referential, Challenge 2 is only partially addressed, as the current token can access only a subset of tokens due to the issue discussed in Challenge 1.

	To address Challenge 1, previous SSM-based vision models frequently employed multi-scanning routes. Specifically, in the case of ViM~\cite{zhu2024ViM}, the token sequence was subjected to both forward and reverse scanning, enabling each token to access global information.
	Although these multi-scan approaches mitigate the causal property of SSMs, they do not address Challenge 2, as the long-range decay characteristic of SSMs remains confined to a 1D format and does not extend into 2D. 
	To enable the acquisition of global information and thus suit non-causal image data, we also begin with a bidirectional scanning strategy. And we demonstrate that the results of forward and reverse scanning of Eq.~\ref{eq:ssd_linear_rewritten} can be integrated to effectively address the aforementioned two challenges simultaneously.
	Let $\mathbf{H}_i$ denote the hidden state of the $i_{th}$ token in the bidirectional scanning approach, from which we can easily derive:
	\begin{equation}
		\mathbf{H}_i = \sum_{j=1}^{i} \frac{1}{A_j} \mathbf{Z}_j + \sum_{j=-L}^{-i} \frac{1}{A_{-j}} \mathbf{Z}_{-j} = \sum_{j=1}^{L} \frac{1}{A_j} \mathbf{Z}_j + \frac{1}{A_i} \mathbf{Z}_i,\text{where } \mathbf{Z}_j = \mathbf{B}_j x(j).
		\label{eq:nc_ssd_bidirection}
	\end{equation}
	If we consider $\frac{1}{A_i}\mathbf{Z}_i$ in this equation as a bias and omit it, Eq.~\ref{eq:nc_ssd_bidirection} can be further simplified, resulting in all tokens sharing the same hidden state $\textbf{H} = \sum_{j=1}^{L} \frac{1}{A_j} \mathbf{Z}_j$.
	In such cases, the results from forward and reverse scanning can be seamlessly combined to establish a global context, effectively equivalent to removing the causal mask and transitioning to a non-causal format. Consequently, the first challenge associated with the causal property is resolved. 
	Although the above results are derived from a bidirectional scanning approach, it is evident that in this non-causal format, different scanning routes yield consistent outcomes. In other words, designing specific scanning routes to capture global information becomes unnecessary. Furthermore, as demonstrated in Eq.~\ref{eq:nc_ssd_bidirection}, the contribution of different tokens to the current hidden state is no longer related to their spatial distance. Therefore, processing a flattened 2D feature map into a 1D sequence no longer compromises the original structural relationships. Thus, the second challenge is also resolved. Additionally, as the entire computation process can be conducted in parallel, rather than relying on the recurrent computational methods which are previously necessary for SSMs, there exists an improvement in training and inference speeds.
	\begin{figure}
		\centering
		\includegraphics[width=1.0\linewidth]{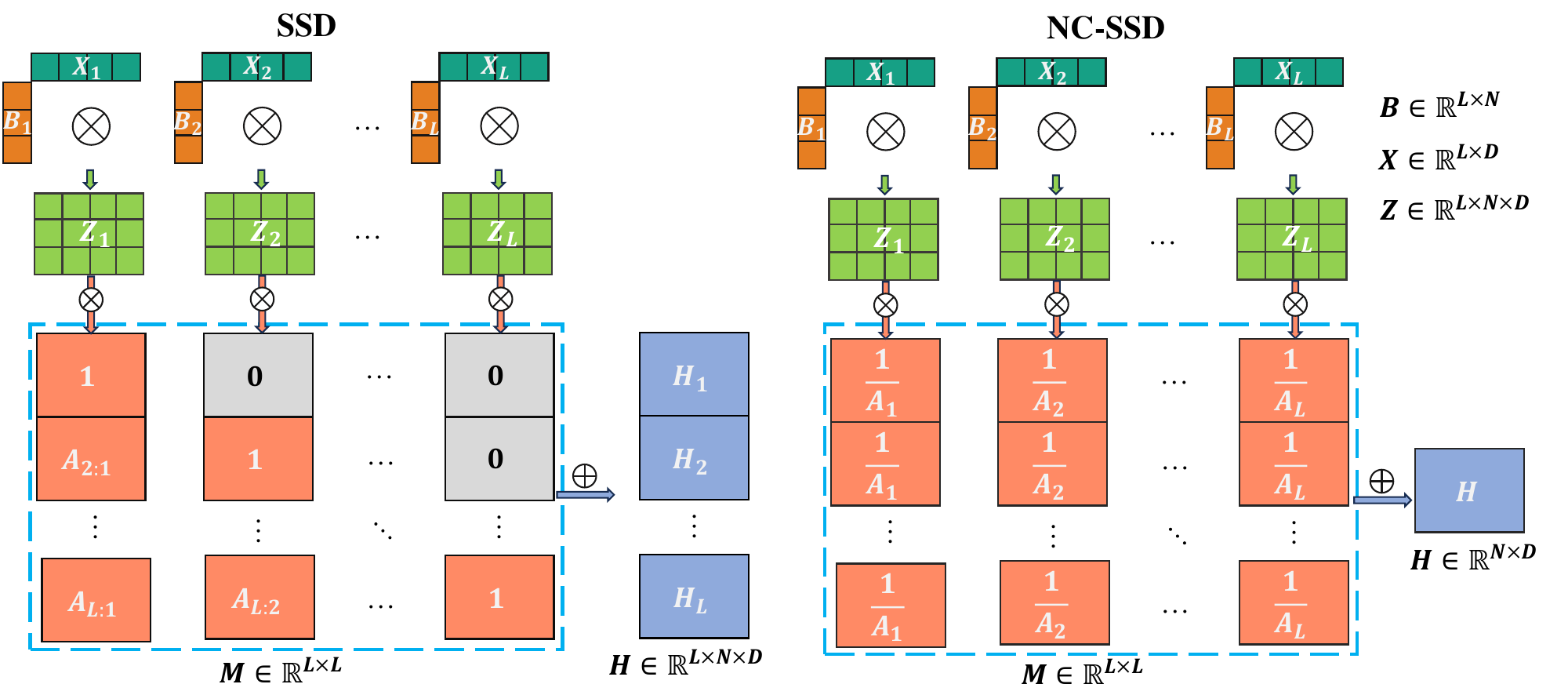}
		\caption{\textbf{Illustration of the Hidden State Generation Process for SSD and NC-SSD}. During the hidden state update process, NC-SSD utilizes the scalar $A$ to determine the extent of information increment for the current token, in contrast to SSD where $A$ dictates the proportion of the hidden state to be retained. Unlike the SSD, which generates token-wise hidden states, the NC-SSD produces only a global hidden state to accommodate non-causal image data.
		}
		\label{fig:ssd}
		\vspace{-5mm}
	\end{figure}
	After revising the iteration rules for the hidden state space, we update the corresponding tensor contraction algorithm or einsum notation in the linear form, following the Mamba2 framework~\cite{dao2024transformers}:
	\begin{equation}
		\begin{aligned}
			\mathbf{Z} & =\operatorname{contract}(\mathrm{LD}, \mathrm{LN} \rightarrow \mathrm{LND})(\mathbf{X}, \mathbf{B}) \\
			\mathbf{H} & =\operatorname{contract}(\mathrm{LL}, \mathrm{LDN} \rightarrow \mathrm{ND})(\mathbf{M}, \mathbf{Z}) \\
			\mathbf{Y} & =\operatorname{contract}(\mathrm{LN}, \mathrm{ND} \rightarrow \mathrm{LD})(\mathbf{C}, \mathbf{H}).
		\end{aligned}
		\label{eq:non_causal_ssd_contraction}
	\end{equation}

	This algorithm involves three steps: the first step expands the input $\mathbf{X}$ using $\mathbf{B}$, the second step unrolls scalar SSM recurrences to create a global hidden state $\mathbf{H}$, and the final step contracts the hidden state $\mathbf{H}$ with $\mathbf{C}$. 
	For clarity, the initial two steps of SSD and NC-SSD are depicted in Fig.~\ref{fig:ssd}. 
	Compared to vallina SSD, while the operation in the first step remains unchanged, the sequence length dimension in the hidden state $\mathbf{H}$ is eliminated in the non-causal mode, as all tokens share the same hidden state. 
	In the final step, the output $\mathbf{Y}$ is produced through the matrix multiplication of $\mathbf{C}$ and $\mathbf{H}$.  Given that $\mathbf{M}_{i,j} = \frac{1}{A_j}$, the matrix $\mathbf{M}$ can be reduced to a vector $\mathbf{m} \in \mathbb{R}^L$ by eliminating its first dimension. In this case, integrating $\mathbf{m}$ with either $\mathbf{X}$ or $\mathbf{B}$ could further simplify the transformation of Eq.~\ref{eq:non_causal_ssd_contraction} to:
	\begin{equation}
		\mathbf{Y} = \mathbf{C}(\mathbf{B}^{T}(\mathbf{X}\cdot\mathbf{m})),
		\label{eq:nc_ssd}
	\end{equation}
	\begin{wrapfigure}{r}{0.6\textwidth}
		\vspace{-3mm}
		\centering
		\includegraphics[width=0.6\textwidth]{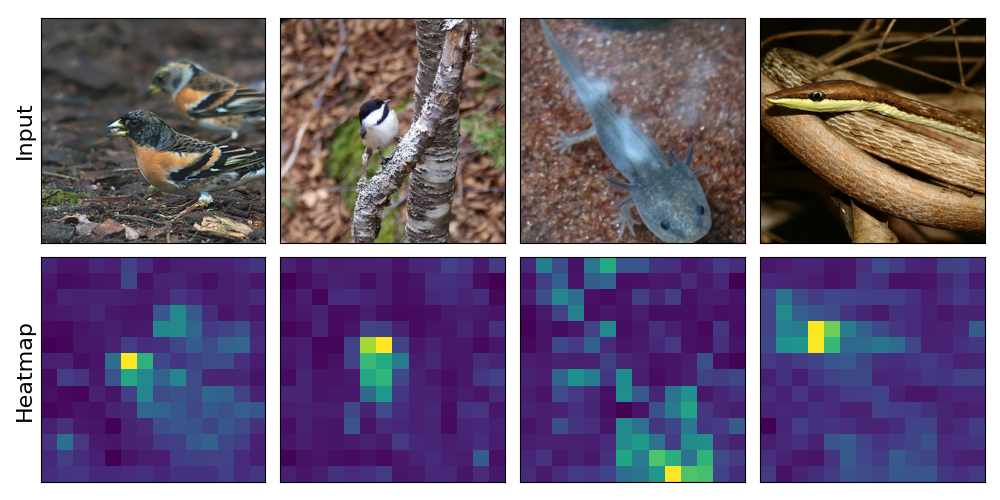}
		\caption{Visualization of input images alongside their corresponding heat maps, which are derived by averaging the vector $\mathbf{m}$ across various heads in the NC-SSD.}
		\label{fig:A_heatmap}
		\vspace{-4mm}
	\end{wrapfigure}
	which can be regarded as a variant of linear attention. 
	However, it is worth noting that just as $A$ plays a distinguished role in Mamba2, the vector $\mathbf{m}$ is also crucial, as demonstrated in our ablation studies.
	In practice, we directly use the learned $A$ instead of $\frac{1}{A}$  since they share the same range of values. 
	To gain a more intuitive understanding of the role of $\mathbf{m}$ in Eq.~\ref{eq:nc_ssd}, we visualize the average of $\mathbf{m}$ across different heads as shown in Fig.~\ref{fig:A_heatmap}. Predominantly, $\mathbf{m}$ focuses on foreground features, enabling the model to prioritize elements that are crucial for the task at hand.

	\subsection{Vision State Space Duality Model}
	\label{sec:vssd}
	\begin{figure}
		\centering
		\includegraphics[width=1.0\linewidth]{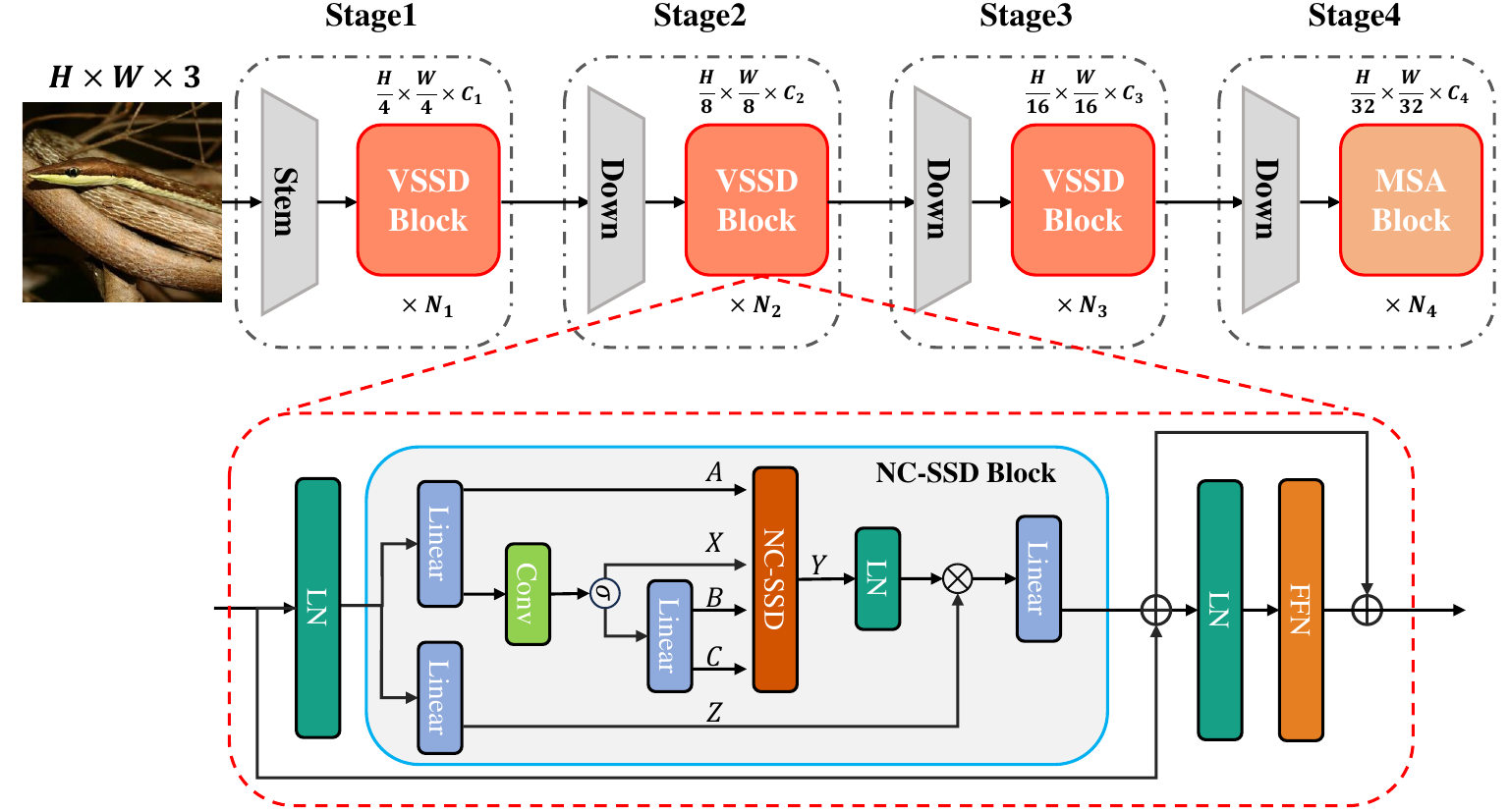}
		\caption{\textbf{Overall Architecture of the Proposed VSSD Model}. The VSSD model initiates with a series of overlapping convolutions serving as the stem, followed by four progressive stages of processing. First three stages are equipped with VSSD Block, which is elaborated in the lower part of the figure, comprising a NC-SSD block and a FFN. Local Perception Units (LPU) are omitted in this visualization for brevity.}
		\label{fig:overall_arc}
		\vspace{-3mm}
	\end{figure}
	\textbf{Block Design.}
	To enhance the SSD block in Mamba2 for vision applications, several modifications have been implemented beyond merely substituting the SSD with an NC-SSD to develop our Visual State Space Duality (VSSD) block. 
	When constructing the NC-SSD block, the causal convolution 1D is replaced by a Depth-Wise Convolution (DWConv) with a kernel size of three, in line with previous vision mamba works~\cite{liu2024vmamba,huang2024localmamba}. 
	Additionally, a Feed-Forward Network (FFN) is integrated subsequent to the NC-SSD block to facilitate enhanced information exchange across channels and to maintain alignment with the established practices of classical vision transformers~\cite{ViT16x16,Swin,DeiT2021}. 
	Moreover, a Local Perception Unit (LPU)~\cite{guo2022cmt} is incorporated prior to the NC-SSD block and the FFN, augmenting the model's capability for local feature perception. 
	Skip connections~\cite{resnet} are also implemented among different blocks. 
	The architecture of the VSSD block is depicted in the lower part of Fig.~\ref{fig:overall_arc}.
	
	\textbf{Hybrid with Self-Attention.}
	Mamba2 demonstrates that integrating SSD with standard Multi-head Self Attention (MSA) yields additional improvements. In a similar vein, our model incorporates self-attention. However, unlike Mamba2, which uniformly intersperses self-attention throughout the network, we strategically replace the NC-SSD block with self-attention module exclusively in the last stage. This modification leverages the robust capabilities of self-attention in processing high-level features, as evidenced by prior works~\cite{lin2023scale,ren2023sg, fan2024rmt} in vision tasks.

	\textbf{Overlapped downsampling layers.} As hierarchical vision transformers~\cite{Swin} and vision state space models~\cite{liu2024vmamba} predominantly employ non-overlapped convolution for downsampling, recent studies~\cite{nat,cvt} have demonstrated that overlapped downsampling convolutions can introduce beneficial inductive biases. Consequently, we adopt overlapped convolutions in our model, following the manner used in MLLA~\cite{han2024demystify}. To maintain the parameter count and computational FLOPs remain comparable, we have accordingly adjusted the depth of our model.

	\textbf{Overall Architecture.}  We develop our VSSD model in accordance with the methods discussed above, and its architecture is depicted in Fig.~\ref{fig:overall_arc}. Mirroring the design principles of established vision backbones in previous works~\cite{Swin,convnext,liu2024vmamba}, our VSSD model is structured into four hierarchical stages. The first three stages employ the VSSD block, while the final stage incorporates the MSA block. Detailed architectures of VSSD variants are shown in the Tab~\ref{tab:model_specs}.
	
	\begin{table}[ht]
		\vspace{-3mm}
		\centering
		\caption{\textbf{Model Specifications of VSSD varints.}}
		\label{tab:model_specs}
		\begin{tabular}{l|ccc|cc}
			\toprule
			\textbf{Model} & \textbf{Blocks} & \textbf{Channels} & \textbf{Heads} & \textbf{\#Param} & \textbf{FLOPs} \\ 
			\midrule
			\textbf{VSSD-M}icro & [2, 2,  \phantom{1}8, 4] & [48, \phantom{1}96, 192, 384] & [2, 4, \phantom{1}8, 16] &  14 & \phantom{1}2.3 \\ 
			\textbf{VSSD-T}iny & [2, 4,  \phantom{1}8, 4] & [64, 128, 256, 512] & [2, 4, \phantom{1}8, 16] &  24 & \phantom{1}4.5 \\ 
			\textbf{VSSD-S}mall & [3, 4, 18, 5] & [64, 128, 256, 512] & [2, 4, \phantom{1}8, 16] & 40 & \phantom{1}7.4 \\ 
			\textbf{VSSD-B}ase & [3, 4, 18, 5] & [96, 192, 384, 768] & [3, 6, 12, 24] & 89 & 16.1 \\
			\bottomrule
		\end{tabular}
		\vspace{-5mm}
	\end{table}
	
	\section{Experiment}
	\subsection{Classification}
	
	\begin{center}
		\captionof{table}{\textbf{Accuracy Comparison across Various Models on ImageNet-1K.} The $\dagger$ indicates results are obtained with MESA~\cite{du2022sharpness}. The LAttn is the abbreviation of linear attention.}
		\label{tab:imagenet}
		\small  
		\setlength{\tabcolsep}{0.03cm} 
		\begin{multicols}{2}
			\begin{tabular}{l|c|cc|c}
				\toprule
				\textbf{Method} & \textbf{Type} &\textbf{\#Param.} & \textbf{FLOPs}  & \begin{tabular}[c]{@{}c@{}} \textbf{Top-1} \\ \textbf{Acc(\%)}\end{tabular} \\
				\midrule
				\multicolumn{5}{c}{\textbf{Micro Models}} \\
				RegNetY-1.6G~\cite{radosavovic2020designing} & Conv & 11M & 1.6G  & 78.0 \\
				EffNet-B3~\cite{efficientnet} &  Conv & 12M & 1.8G  & 81.6 \\
				PVTv2-b1~\cite{pvtv2} & Attn & 13M &2.1G &78.7 \\
				BiFormer~\cite{biformer} & Attn & 13M &2.2G &81.4\\
				NAT-M~\cite{nat} & Attn & 20M &2.7G &81.8\\
				CMT-XS~\cite{guo2022cmt}& Attn & 15M &1.5G &81.8\\
				SMT-T~\cite{lin2023scale} & Attn & 12M &2.4G &82.2\\
				Vim-T~\cite{zhu2024ViM} & SSM & 7M &1.5G &76.1\\
				LVim-T~\cite{huang2024localmamba} & SSM & 8M &1.5G &76.2\\
				\grayrow VSSD-M & SSD & 14M &2.3G &\textbf{82.5}\\
				\midrule
				\multicolumn{5}{c}{\textbf{Tiny Models}} \\
				RegNetY-4G~\cite{radosavovic2020designing} &Conv& 21M & 4.0G  & 80.0\\ 
				ConvNeXt-T~\cite{convnext} &Conv& 29M & 4.5G  & 82.1\\
				MambaOut-T~\cite{yu2024mambaout} &Conv& 27M & 4.5G  & 82.7\\
				EffNet-B4~\cite{efficientnet}&Conv& 19M & 4.2G  & 82.9\\
				DeiT-S~\cite{DeiT2021} &Attn& 22M & 4.6G  & 79.8 \\
				Swin-T~\cite{Swin}&Attn& 29M & 4.5G  & 81.3\\
				PVTv2-B2~\cite{pvtv2}&Attn& 25M & 4.0G  & 82.0\\
				Focal-T~\cite{yang2022focal}&Attn& 29M & 4.9G  & 82.2\\
				CSwin-T~\cite{cswin}&Attn& 23M & 4.3G  & 82.7\\
				NAT-T~\cite{nat}&Attn& 28M & 4.3G  & 83.2\\
				VMambaV9-T~\cite{liu2024vmamba} & SSM & 31M &4.9G &82.5\\
				LVMamba-T~\cite{huang2024localmamba} & SSM & 26M &5.7G &82.7\\
				MSVMamba-T~\cite{shi2024multi}& SSM & 33M &4.6G &82.8\\
				\grayrow VSSD-T& SSD & 24M &4.5G &\textbf{83.7}\\
				MLLA-T~\cite{han2024demystify} & LAttn & 25M & 4.2G &83.5$^{\dagger}$\\
				\grayrow VSSD-T& SSD & 24M &4.5G &\textbf{84.1}$^{\dagger}$\\
				\bottomrule
			\end{tabular}
			
			\columnbreak
			\begin{tabular}{l|c|cc|c}
				\toprule
				\textbf{Method} & \textbf{Type} &\textbf{\#Param.} & \textbf{FLOPs}  & \begin{tabular}[c]{@{}c@{}} \textbf{Top-1} \\ \textbf{Acc(\%)}\end{tabular} \\
				\midrule
				
				\multicolumn{5}{c}{\textbf{Small Models}} \\
				ConvNeXt-S~\cite{convnext} &Conv& 50M & 8.7G  & 83.1\\
				EffNet-B5~\cite{efficientnet}&Conv& 30M & 9.9G  & 83.6\\
				MambaOut-S~\cite{yu2024mambaout} &Conv& 48M & 9.0G  & 84.1\\
				Swin-S~\cite{Swin} &Attn& 50M & 8.7G  & 83.0 \\
				PVTv2-B3~\cite{pvtv2}&Attn& 45M & 6.9G  & 83.2\\
				Focal-S~\cite{yang2022focal}&Attn& 50M & 8.7G  & 83.5\\
				CSwin-S~\cite{cswin}&Attn& 35M & 6.9G  & 83.6\\
				NAT-S~\cite{nat}&Attn& 51M & 7.8G  & 83.7\\
				VMamba-S~\cite{liu2024vmamba} & SSM & 44M &11.2G &83.5\\
				PMamba-L2~\cite{yang2024plainmamba} & SSM & 25M &8.1G &81.6\\
				VMambaV9-S~\cite{liu2024vmamba} & SSM & 50M &8.7G &83.6\\
				LVMamba-S~\cite{huang2024localmamba} & SSM & 50M &11.4G &83.7\\
				\grayrow VSSD-S& SSD & 40M &7.4G &\textbf{84.1}\\
				MLLA-S~\cite{han2024demystify} & LAttn & 43M & 7.3G &84.4$^{\dagger}$\\
				\grayrow VSSD-S& SSD & 40M &7.4G &\textbf{84.5}$^{\dagger}$\\
				\midrule
				\multicolumn{5}{c}{\textbf{Base Models}} \\
				ConvNeXt-B~\cite{convnext} &Conv& 89M & 15.4G  & 83.8\\
				MambaOut-B~\cite{yu2024mambaout} &Conv& 85M & 15.8G  & 84.2\\
				DeiT-B~\cite{DeiT2021} &Attn& 86M & 17.5G  & 81.8 \\
				Swin-B~\cite{Swin} &Attn& 88M & 15.4G  & 83.5 \\
				CSwin-S~\cite{cswin}&Attn& 78M & 15.0G  & 84.2\\
				NAT-B~\cite{nat}&Attn& 90M & 13.7G  & 84.3\\
				PMamba-L3~\cite{yang2024plainmamba} & SSM & 50M &14.4G &82.3\\
				VMambaV9-B~\cite{liu2024vmamba} & SSM & 89M &15.4G &83.9\\
				\grayrow VSSD-B& SSD & 89M &16.1G &\textbf{84.7}\\
				MLLA-B~\cite{han2024demystify} & LAttn & 96M & 16.2G &85.3$^{\dagger}$\\
				\grayrow VSSD-B& SSD & 89M &16.1G &\textbf{85.4}$^{\dagger}$\\
				\bottomrule
			\end{tabular}
		\end{multicols}
	\end{center}
	
	\textbf{Configurations}. Our experiments are conducted using the ImageNet-1K dataset~\cite{deng2009imagenet}, consistent with methodologies from prior studies~\cite{Swin,liu2024vmamba}. Each model is subjected to a training regimen spanning 300 epochs, which includes a 20-epoch warm-up phase. 
	Optimization is performed with AdamW, where the betas are set to (0.9, 0.999) and momentum at 0.9. A cosine decay scheduler manages the learning rate and is combined with a weight decay rate of 0.05. To further refine model accuracy and generalization, we incorporate exponential moving average (EMA) techniques and apply label smoothing with a coefficient of 0.1.  More detailed configurations could be found in Appendix~\ref{sec:more_details}.
	For the testing phase, images are center-cropped to dimensions of 224$\times$224.

	\textbf{Performance Evaluation.} Tab.~\ref{tab:imagenet} presents a comparative comparison of our VSSD models against CNNs, ViTs, and other SSM-based frameworks on the ImageNet-1K dataset~\cite{deng2009imagenet}. The VSSD-M model, equipped with 14M parameters and 2.3G FLOPs, secures a top-1 accuracy of 82.5\%, surpassing the similarly priced NAT-M~\cite{nat} by 0.7\%. In comparisons with models categorized as tiny and small, VSSD consistently outperforms its counterparts. Specifically, the VSSD-T model, which has 24M parameters and 4.5G FLOPs, achieves an accuracy of 83.7\%, outdoing VMambaV9-T~\cite{liu2024vmamba} by 1.2\%. For the small-sized model variant, VSSD-S, which comprises 40M parameters and 7.4G FLOPs, achieves an accuracy of 84.1\%, surpassing the LocalVMamba-S~\cite{huang2024localmamba} by 0.4\%. In the base-sized variants, our VSSD-B, with 89M parameters and 16.1G FLOPs, records an accuracy of 84.7\%, which is 0.8\% higher than that of VMambaV9-B. When introducing MESA~\cite{du2022sharpness} for further refinement, the results for our tiny, small and base size model are improved to 84.1\%, 84.5\% and 85.4\% respectively.

	\begin{figure}
		\centering
		\includegraphics[width=1.0\linewidth]{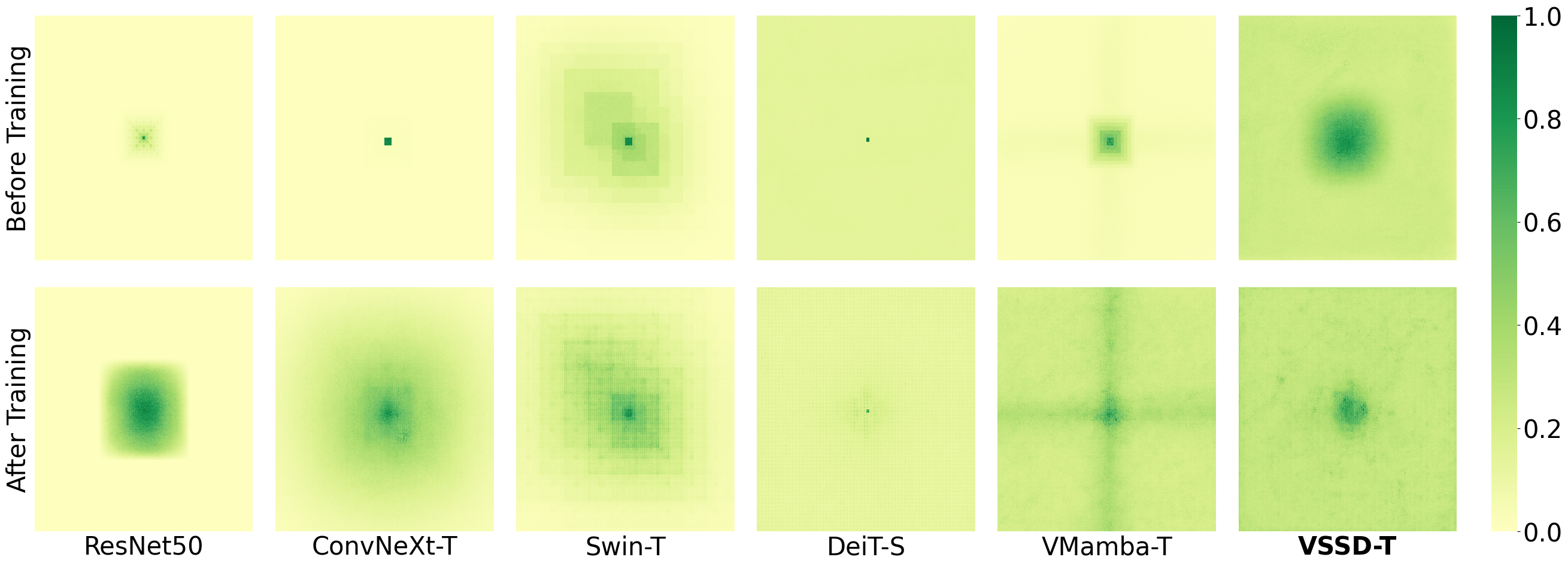}
		\caption{ \textbf{Comparison of the Effective Receptive Field (ERF)} among our VSSD, CNN-based models (ResNet~\cite{resnet} and ConvNeXt~\cite{convnext}), attention-based models (Swin~\cite{Swin} and DeiT~\cite{DeiT2021}), and the SSM-based VMamba~\cite{liu2024vmamba}. Our VSSD effectively eliminates the impact of token spacing on the contribution of information compared to SSM-based VMamba.}
		\label{fig:erf}
		\vspace{-3mm}
	\end{figure}
	In addition to quantitative comparisons, we conducted a comparative analysis of the Effective Receptive Field (ERF) before and after training across various models, including CNN-based ResNet50~\cite{resnet} and ConvNeXt-Tiny~\cite{convnext}, attention-based Swin-Tiny~\cite{Swin} and DeiT-Small~\cite{DeiT2021}, and SSM-based VMamba-Tiny~\cite{liu2024vmamba}, along with our VSSD-Tiny.
	The ERF of the central pixel is plotted using the method proposed in ~\cite{luo2016understanding}, utilizing 50 randomly selected images with a resolution of 1024x1024 from the ImageNet-1K validation set.
	To demonstrate the effectiveness of the proposed NC-SSD, the techniques such as hybrid self-attention and overlapped downsampling layers discussed in Sec.~\ref{sec:vssd} are not employed in our VSSD model for this analysis.
	Notably, only our VSSD and DeiT consistently exhibit a global receptive field both before and after training. Under after-training, a distinct cross-shaped attenuation is observed in VMamba, whereas our approach effectively eliminates the impact of token spacing on the contribution of information.

	\subsection{Object Detection and Instance Segmentation}

	\textbf{Configurations.} Our evaluation of the VSSD model utilizes the MS COCO dataset~\cite{coco} within the Mask R-CNN framework~\cite{maskrcnn} for tasks related to object detection and instance segmentation. All experiments are facilitated using the MMDetection library~\cite{chen2019mmdetection}. Consistent with prior studies~\cite{Swin,liu2024vmamba}, during the training phase, images are adjusted such that the shorter side measures 800 pixels, while the longer side does not exceed 1333 pixels. Optimization is carried out using the AdamW optimizer, with a set learning rate of 0.0001 and a batch size of 16. When adopting the standard "1×" training schedule, the learning rate is reduced by a factor of 0.1 at epochs 8 and 11 while the extended "3× + MS" schedule sees a reduction in the learning rate by the same factor at epochs 27 and 33.

	\textbf{Performance Evaluation.} Tab.~\ref{table:coco} details the comparative performance of our model against well-established CNNs, ViTs, and other SSM-based models. Our VSSD model demonstrates superior performance in various configurations. Remarkably, our VSSD-T model demonstrates a significant advantage, outperforming Swin-T~\cite{Swin} by margins of +4.2 in box AP and +3.3 in mask AP.
	Under the extended "3×" training schedule, VSSD-T still consistently outperforms various competitors.
	
	\begin{table}
		\centering
		\caption{\textbf{Object Detection and Instance Segmentation with Mask R-CNN Framework~\cite{maskrcnn} on MS COCO Dataset~\cite{coco}}.  The FLOPs are tested with an input size of $1280 \times 800$. }
		\label{table:coco}

		\begin{tabular}{l|cccccc|cc}
			\toprule
			\multicolumn{9}{c}{\textbf{Mask R-CNN 1x}} \\
			\textbf{Method} &AP$^\text{b}$ & AP$^\text{b}_\text{50}$ & AP$^\text{b}_\text{75}$ & AP$^\text{m}$ & AP$^\text{m}_\text{50}$ & AP$^\text{m}_\text{75}$ & \textbf{\#Param.} & \textbf{FLOPs} \\
			\midrule
			PVT-T~\cite{PVT} & 36.7 & 59.2 & 39.3 & 35.1 & 56.7 & 37.3 & 33M & 208G \\
			EffVMamba-S~\cite{pei2024efficientvmamba} & 39.3 & 61.8 & 42.8 & 36.7 & 58.9 & 39.2 & 31M & 197G \\
			MSVMamba-M~\cite{shi2024multi} & 43.8 & 65.8 & 47.7 & 39.9 & 62.9 & 42.9 & 32M & 201G \\
			\grayrow VSSD-M & 45.4 & 67.5 & 49.8 & 41.3 & 64.5 & 44.6 & 33M & 220G \\
			\midrule
			Swin-T~\cite{Swin} & 42.7 & 65.2 & 46.8 & 39.3 & 62.2 & 42.2 & 48M & 267G \\
			ConvNeXt-T~\cite{convnext} & 44.2 & 66.6 & 48.3 & 40.1 & 63.3 & 42.8 & 48M & 262G \\
			VMamba-T~\cite{liu2024vmamba} & 46.5 & 68.5 & 50.7 & 42.1 & 65.5 & 45.3 & 42M & 286G \\
			LocalVMamba-T~\cite{huang2024localmamba} & 46.7 & 68.7 & 50.8 & 42.2 & 65.7 & 45.5 & 45M & 291G \\
			MSVMamba-T~\cite{shi2024multi} & 46.9 & 68.8 & 51.4 & 42.2 & 65.6 & 45.4 & 53M & 252G \\
			\grayrow VSSD-T & 46.9 & 69.4 & 51.4 & 42.6 & 66.4 & 45.9 & 44M & 265G \\
			\midrule
			Swin-S~\cite{Swin} & 44.8 &66.6& 48.9& 40.9& 63.2& 44.2 & 69M & 354G \\
			ConvNeXt-S~\cite{convnext} & 45.4 & 67.9 & 50.0 & 41.8 & 65.2 & 45.1 & 70M & 348G \\
			VMamba-S~\cite{liu2024vmamba} & 48.2& 69.7& 52.5& 43.0& 66.6& 46.4 & 64M & 400G \\
			LocalVMamba-S~\cite{huang2024localmamba} & 48.4& 69.9& 52.7& 43.2 &66.7 &46.5 & 69M & 414G \\
			\grayrow VSSD-S & 48.4 & 70.1 & 53.1 & 43.5 & 67.2 & 47.1 & 59M & 325G \\
			\bottomrule
			\multicolumn{9}{c}{\textbf{Mask R-CNN 3x + MS}} \\
			\textbf{Method} &AP$^\text{b}$ & AP$^\text{b}_\text{50}$ & AP$^\text{b}_\text{75}$ & AP$^\text{m}$ & AP$^\text{m}_\text{50}$ & AP$^\text{m}_\text{75}$ & \textbf{\#Param.} & \textbf{FLOPs} \\
			\midrule
			PVT-T~\cite{PVT} & 39.8 & 62.2 & 43.0 & 37.4 & 59.3 & 39.9 & 33M & 208G \\
			LightViT-T~\cite{huang2022lightvit}& 41.5 & 64.4 & 45.1 & 38.4 & 61.2 & 40.8 & 28M & 187G \\
			EffVMamba-S~\cite{pei2024efficientvmamba} & 41.6 & 63.9 & 45.6 & 38.2 & 60.8 & 40.7 & 31M & 197G \\
			MSVMamba-M~\cite{shi2024multi} & 46.3 & 68.1 & 50.8 & 41.8 & 65.1 & 44.9 & 32M & 201G \\
			\grayrow VSSD-M & 47.7 & 69.7 & 52.1 & 42.8 & 66.5 & 46.0 & 33M & 220G \\
			\midrule
			Swin-T~\cite{Swin} & 46.0 & 68.1 & 50.3 & 41.6 & 65.1 & 44.9 & 48M & 267G \\
			ConvNeXt-T~\cite{convnext} & 46.2 & 67.9 & 50.8 & 41.7 & 65.0 & 44.9 & 48M & 262G \\
			VMamba-T~\cite{liu2024vmamba} & 48.5 & 69.9 & 52.9 & 43.2 & 66.8 & 46.3 & 42M & 286G \\
			LocalVMamba-T~\cite{huang2024localmamba} & 48.7 & 70.1 & 53.0 & 43.4 & 67.0 & 46.4 & 45M & 291G \\
			\grayrow VSSD-T & 48.8 & 70.4 & 53.4 & 43.6 & 67.6 & 46.9 & 44M & 265G \\
			\bottomrule
		\end{tabular}

		\vspace{-2mm}
	\end{table}
	
	\subsection{Semantic Segmentation}
	
	\textbf{Configurations.} In alignment with the approaches described in Swin~\cite{Swin} and VMamba~\cite{liu2024vmamba}, our experiments leverage the UperHead~\cite{upernet} framework, utilizing an ImageNet pre-trained backbone for initialization. The training regimen spans 160K iterations with a batch size of 16, executed using the MMSegmentation library~\cite{mmseg2020}. The primary experiments are performed with a standard input resolution of $512 \times 512$. To further evaluate the robustness of our model, Multi-Scale (MS) testing is implemented. Optimization is carried out using the AdamW optimizer, with the learning rate established at $6 \times 10^{-5}$.
	
	\textbf{Performance Evaluation.} Detailed performance metrics for our model and its competitors are displayed in Tab.~\ref{table:ade20k}, encompassing both single-scale and multi-scale testing scenarios. Specifically, in the context of the Tiny model category and single-scale testing, our VSSD model demonstrates superior performance, exceeding the results of Swin, ConNeXt, and VMamba models in the tiny variant by margins of +3.5, +1.9, and +0.6 mIoU, respectively.

	\begin{table}
		\centering
		\caption{\textbf{Results of Semantic Segmentation on the ADE20K Dataset~\cite{ade20k} using the UperNet Framework~\cite{upernet}}. FLOPs for all models are computed using input dimensions of $512 \times 2048$. In the table, "SS" represents single-scale testing, while "MS" indicates multi-scale testing.}
		\label{table:ade20k}
		\begin{tabular}{l|cc|cc}
			\toprule
			\textbf{Method} &  \begin{tabular}[c]{@{}c@{}} \textbf{mIoU} \\ \textbf{SS} \end{tabular} & \begin{tabular}[c]{@{}c@{}} \textbf{mIoU} \\ \textbf{MS} \end{tabular} & \textbf{\#Param}. & \textbf{FLOPs} \\
			\midrule
			EffVMamba-S~\cite{pei2024efficientvmamba} & 41.5 & 42.1 & 29M & 505G  \\
			MSVMamba-M~\cite{shi2024multi} & 45.1 & 45.4 & 42M & 875G  \\
			\grayrow VSSD-M & 45.6 & 46.0 & 42M & 893G  \\
			\midrule
			Swin-T~\cite{Swin} & 44.4 & 45.8 & 60M & 945G \\
			ConvNeXt-T~\cite{convnext} & 46.0 & 46.7 & 60M & 939G \\
			VMamba-T~\cite{liu2024vmamba} & 47.3 & 48.3 & 55M & 964G  \\
			LocalVMamba-T~\cite{huang2024localmamba} & 47.9 & 49.1 & 57M & 970G  \\
			EffVMamba-B~\cite{pei2024efficientvmamba} & 46.5 & 47.3 & 65M & 930G  \\
			MSVMamba-T~\cite{shi2024multi} & 47.6 & 48.5 & 65M & 942G  \\
			
			\grayrow VSSD-T & 47.9 & 48.7 & 53M & 941G  \\
			\bottomrule
		\end{tabular}
	\end{table}

\subsection{Ablations}

To validate the effectiveness of the proposed modules, we conducted detailed ablation experiments on the VSSD-Micro model. Using the SSD block as the token mixer and patchified downsamplers (e.g. convolution with $4\times4$ kernel and stride of $4$ in stem) following Swin~\cite{Swin} and vallina VMamaba~\cite{liu2024vmamba}, we established the baseline configuration, detailed in the first row of Tab.~\ref{tab:model_ablation}. For throughput testing, we utilized an A100-PCIE-40G GPU with a batch size of 128 and FP16 precision.

\setlength{\tabcolsep}{1.5pt}
\begin{table}[!ht]
	\centering
	\caption{
		\textbf{Ablation study of VSSD-Micro on ImageNet-1K.} Our NC-SSD consistently outperforms vallina SSD and Bi-SSD in terms of accuracy and efficiency. Other techniques further enhance the performance.
	}
	\label{tab:model_ablation}
	\begin{tabular}{lcccc|cccc}
		\toprule
		\textbf{Op. Type} & \textbf{Downsampler} & \textbf{Layers}     & \textbf{Top-1} & \textbf{\#Params}   & \textbf{FLOPs} & \textbf{Thru.} & \textbf{Train Thru.}\\
		&  &     &\textbf{Acc}(\%)           &   & (G) & (imgs/sec) & (imgs/sec) & \\
		\midrule
		\textbf{SSD}       & Patch &  2, 4, 8, 4 & 81.0 & 14.8 M & 2.1 & 1818 & 523 \\
		\midrule
		\textbf{Bi-SSD}    & Patch  & 2, 4, 8, 4 & 81.4 & 15.2 M & 2.2 & 1741 & 399  \\
		\textbf{NC-SSD}    & Patch  & 2, 4, 8, 4 & 81.6 & 14.8 M & 2.1 & 1843 & 606  \\
		\textbf{Hybrid}    & Patch   & 2, 4, 8, 4  & 81.8 & 13.4 M & 2.1 & 1890 & 622  \\
		\grayrow \textbf{Hybrid}    & Conv   & 2, 2, 8, 4 & 82.5 & 13.5 M & 2.3 & 1918 & 597 \\
		\bottomrule
	\end{tabular}
	
\end{table}

\textbf{Different SSD Mechanisms.} In our ablation study for the token mixer, we explored different scanning routes for SSD. Specifically, the Bi-SSD is introduced, where we split the channels into two parts and reverse one part to create backward scanning sequences. These sequences with opposite scanning routes are then concatenated after the SSD block. As shown in Tab.~\ref{tab:model_ablation}, our NC-SSD model outperforms both the vanilla SSD and Bi-SSD by 0.6\% and 0.2\% in top-1 accuracy, respectively. Moreover, both training and inference throughput are enhanced, with NC-SSD improving training throughput by nearly 50\% compared to the Bi-SSD approach.

\textbf{Hybrid Architecture and Overlapped Downsampler.} The effectiveness of incorporating standard attention in the last stage and using overlapped downsampler is demonstrated in the last two rows of Tab.~\ref{tab:model_ablation}. Specifically, replacing NC-SSD with standard attention in the last stage results in a 0.2\% improvement in accuracy while slightly reducing the parameters. Replacing the patchified downsampler with the overlapped convolutional manner improves accuracy by 0.7\% while increasing the FLOPs by 0.2G. To maintain approximate parameters, we adjusted the layer configuration from [2,4,8,4] to [2,2,8,4].


\textbf{Effect of $\mathbf{m}$.} Eq.~\ref{eq:nc_ssd} conceptualizes NC-SSD as a variant of linear attention that incorporates an additional weight vector, $\mathbf{m}$. Fig.~\ref{fig:A_heatmap} visually demonstrates how $\mathbf{m}$ selectively emphasizes foreground features. To quantitatively assess the impact of $\mathbf{m}$, we conducted experiments with and without this component in the NC-SSD block under 100 training epochs with 5 epochs for warming up. The results, presented in Tab.~\ref{tab:m_ablation}, reveal a significant influence of $\mathbf{m}$ on model performance. 

\begin{wraptable}{r}{0.5\textwidth}
	\centering
	\caption{\textbf{Ablation Study on the Effect of $\mathbf{m}$ in NC-SSD.} The symbol $\dagger$ indicates the best accuracy achieved prior to encountering \textbf{N.A}.}
	\label{tab:m_ablation}
	\begin{tabular}{lc|c|ccc}
		\toprule
		\textbf{Operation} & \textbf{Size}  & $\mathbf{m}$     & \textbf{Top-1} & \textbf{\#Params}   & \textbf{FLOPs}\\
		& &                &  \textbf{Acc}(\%)           &   & (G)\\
		\midrule
		\multirow{3}{*}{\textbf{NC-SSD}} & \multirow{2}{*}{\textbf{Tiny}}  & \xmark & 32.6$^{\dagger}$ & 24.3M & 4.5 \\
		& &\checkmark & 81.8 & 24.3M & 4.5 \\
		\cmidrule{2-6}
		& \textbf{Small} & \xmark  &  \textbf{N.A} & 40.0M & 7.4 \\
		\bottomrule
	\end{tabular}
\end{wraptable}
Without $\mathbf{m}$, our experiments show that the model experiences unstable training, leading to crashes. This instability is particularly pronounced in larger models. We report the highest accuracy achieved before training crashed, marked with a $\dagger$. For the tiny-sized model, the best accuracy was only 32.6\%. For the small-sized model, the training crashed in the very first epoch.
We hypothesize that this instability arises because, in the absence of normalization techniques typically used in linear attention approaches~\cite{choromanski2020rethinking,qin2022cosformer,shen2021efficient}, the magnitude of the features escalates dramatically, leading to a crash. In contrast, with $\mathbf{m}$, the model achieves a robust top-1 accuracy of 81.8\%, while maintaining the same number of parameters and computational complexity.

\section{Limitations}
Although the proposed VSSD model outperforms other SSM-based models in ImageNet-1K, the incremental performance gains of VSSD on downstream tasks~\cite{coco,ade20k}, when compared to other SSM-based models, are marginal. When evaluated against SOTA vision transformer variants~\cite{fan2024rmt,xia2023dat++,shi2024transnext}, there remains a significant gap in performance on downstream tasks. 
Furthermore, this paper lacks experiments involving larger models and more extensive datasets, such as those using the ImageNet-22K benchmark~\cite{deng2009imagenet}. Consequently, the scalability of the proposed VSSD model remains an area ripe for further exploration.

\section{Conclusion}
In conclusion, our study introduces the NC-SSD, which redefines SSD by modifying the role of matrix $\mathbf{A}$ and eliminating the causal mask. These adaptations facilitate a transition to a non-causal mode, significantly enhancing both accuracy and efficiency. Extensive experiments demonstrate its superiority over the vanilla SSD and its multi-scan based variants.
Furthermore, by integrating techniques such as hybrid standard attention and overlapped downsampling, our VSSD model achieves comparable or superior performance compared to well-established CNNs, ViTs, and Vision SSMs across several widely-used benchmarks.

\clearpage
\bibliographystyle{plain}
\bibliography{neurips_2024}

\clearpage
\appendix
\section{More Detailed information of VSSD}
\label{sec:more_details}
Our experiments are conducted using the ImageNet-1K dataset~\cite{deng2009imagenet}. Each model undergoes training for 300 epochs, which includes a 20-epoch warm-up phase. We employ the AdamW optimizer, setting the betas to (0.9, 0.999) and the momentum to 0.9. A cosine decay scheduler manages the learning rate, complemented by a weight decay rate of 0.05. The batch sizes and peak learning rates are set to 1024/1e-3 for the Micro and Tiny models, and 2048/1.2e-3 for the Small and Base models, respectively. To enhance model accuracy and generalization, we incorporate exponential moving average (EMA) techniques and apply label smoothing with a coefficient of 0.1. The stochastic depth drop rates for our Micro, Tiny, Small, and Base models are set at 0.2, 0.2, 0.4, and 0.6, respectively. Further details are provided in Tab.~\ref{tab:vssd_config}. 
\begin{table}[ht]
	\centering
	\caption{\textbf{Detailed Configuration Parameters for ImageNet-1K Training.}}
	\label{tab:vssd_config}
	\begin{tabular}{@{}l|cccc@{}}
		\toprule
		\textbf{Settings}       & \textbf{Micro} & \textbf{Tiny} & \textbf{Small} & \textbf{Base} \\ \midrule
		Input resolution          & \multicolumn{4}{c}{$224^2$} \\
		Epochs                    & \multicolumn{4}{c}{300} \\
		Batch size                & 1024 & 1024 & 2048 & 2048 \\
		Optimizer                 & \multicolumn{4}{c}{AdamW} \\
		Adam $\epsilon$           & \multicolumn{4}{c}{1e-8} \\
		Adam $(\beta_1, \beta_2)$ & \multicolumn{4}{c}{(0.9, 0.999)} \\
		Learning rate             & 1e-3 & 1e-3 & 1.2e-3 & 1.2e-3 \\
		Learning rate decay       & \multicolumn{4}{c}{Cosine} \\
		Warmup epochs             & \multicolumn{4}{c}{20} \\
		Weight decay              & \multicolumn{4}{c}{0.05} \\
		Rand Augment              & \multicolumn{4}{c}{rand-m9-mstd0.5-inc1} \\
		Cutmix                    & \multicolumn{4}{c}{1.0} \\
		Mixup                     & \multicolumn{4}{c}{0.8} \\
		Cutmix-Mixup switch prob  & \multicolumn{4}{c}{0.5} \\
		Random erasing prob       & \multicolumn{4}{c}{0.25} \\
		Label smoothing           & \multicolumn{4}{c}{0.1} \\
		Stochastic depth rate& 0.2          & 0.2           & 0.4            & 0.6 \\
		Random erasing prob       & \multicolumn{4}{c}{0.25} \\
		EMA decay rate            & \multicolumn{4}{c}{0.9999} \\ \bottomrule
	\end{tabular}
\end{table}

\end{document}